\documentclass{article}
\usepackage{iclr2027_conference,times}
\usepackage[utf8]{inputenc}
\usepackage[T1]{fontenc}
\usepackage{hyperref}
\usepackage{url}
\usepackage{booktabs}
\usepackage{amsfonts}
\usepackage{amsmath}
\usepackage{nicefrac}
\usepackage{microtype}
\usepackage{xcolor}
\usepackage{graphicx}
\graphicspath{{./}}
\usepackage{multirow}
\usepackage{subcaption}
\usepackage{float}
\usepackage{amssymb}

\title{INTRA: Interleaved Non-contiguous Token spaRse Attention}
\author{Yuxuan Xia \quad Wenbo Lu \quad Shaoyi Zheng \quad Shengjie Wang\\
New York University}
\hypersetup{
  hidelinks,
  pdftitle={INTRA: Interleaved Non-contiguous Token spaRse Attention},
  pdfauthor={Yuxuan Xia, Wenbo Lu, Shaoyi Zheng, Shengjie Wang},
  pdfsubject={Structured sparse attention for efficient Transformer inference}
}
\iclrfinalcopy
\begin{document}
\maketitle
\lhead{Preprint}
\rhead{}
\renewcommand{\headrulewidth}{0pt}

\begin{abstract}

Transformers achieve strong performance across modalities but are bottlenecked by the quadratic cost of attention. Sparse attention can reduce this cost, but effective sparsity must preserve useful long-range communication and map well to GPU hardware. We introduce \textbf{INTRA}--\textbf{I}nterleaved \textbf{N}on-contiguous \textbf{T}oken spa\textbf{R}se \textbf{A}ttention--a hardware-aware static sparse attention design based on a relay-communication view under Computational Query Set (CQS) constraints. Rather than directly covering all token-to-token interactions within each sparse layer, INTRA uses complementary sparse groups to provide short mediated communication paths across layers. Scatter forms relay groups that sample broadly across local neighborhoods, while Gather reconnects these groups in subsequent layers. Under a partition model, this structure guides CQS-compatible sparse patterns with global reachability. Our CQS-aware FlashAttention-style kernel computes structured indices at runtime, loads selected non-contiguous tokens, and packs them densely for computation. On an A100, INTRA achieves a $1.53\times$ speedup in $2\mathrm{K}$ \texttt{FLUX.1-dev} generation, with better measured image quality than evaluated CLEAR configurations after LoRA self-distillation. On LLaMA-3.1-8B, it improves long-context prefill efficiency with strong teacher-forced retrieval performance under matched LoRA adaptation. Attention-only benchmarks further show up to $8.4\times$ and $10.4\times$ speedups over MoBA and density-constrained NSA, respectively. Code is available at \url{https://github.com/NovTi/INTRA}.

\end{abstract}

\section{Introduction}
Transformers excel across domains such as NLP and image generation, but their attention cost scales quadratically with sequence length, creating a bottleneck for long contexts and high-resolution generation. For instance, \texttt{FLUX.1-dev}~\citep{flux2024} requires nearly a minute to produce a 2048$\times$2048 image on a single A100, precluding real-time generation.

Sparse attention reduces full-attention overhead~\citep{liu2024clear,jiang2024minference}. We distinguish input-dependent (\textbf{dynamic}) from input-independent (\textbf{static}) connectivity separately from whether a method requires training or adaptation. INTRA uses static connectivity to avoid runtime pattern selection and evaluates adaptation of pretrained models to GPU-friendly sparse patterns.

Sparsity changes communication as well as FLOPs. Many modalities have coherent local neighborhoods, such as image patches, spatial regions, or short contiguous token segments. Local attention can mix nearby information within these neighborhoods; the remaining challenge is communication between neighborhoods under hardware constraints. A partition-based sparse layer permits interactions only within groups. Covering all dense edges would require many such layers, motivating relay paths that connect neighborhoods through successive groups.

\begin{figure}[h]
  \centering
  \includegraphics[width=1\textwidth]{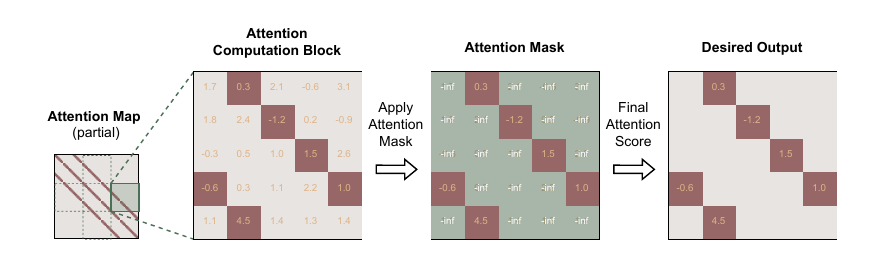}
  \caption{Low-level implementation of blockwise sparsity for handling mixed blocks. A mix block is the block that partially preserves attention scores, as shown on the left side of the figure.}
  \label{fig:mix-blocks}
\end{figure}

Efficient execution constrains this design space. CLEAR~\citep{liu2024clear} uses circular local sparsity, and NATTEN~\citep{hassani2023natten} uses sliding windows. Fine-grained masks can leave partially used computational blocks (Figure~\ref{fig:mix-blocks}). Computing whole blocks before masking unused scores diminishes realized speedups relative to nominal FLOP reductions. We formalize a hardware-oriented design class through Computational Query Sets (CQSs): query groups with structured key sets and local-to-global index mappings aligned with FlashAttention-style query blocks.
\begin{figure}[h]
    \centering
    \includegraphics[width=1\linewidth]{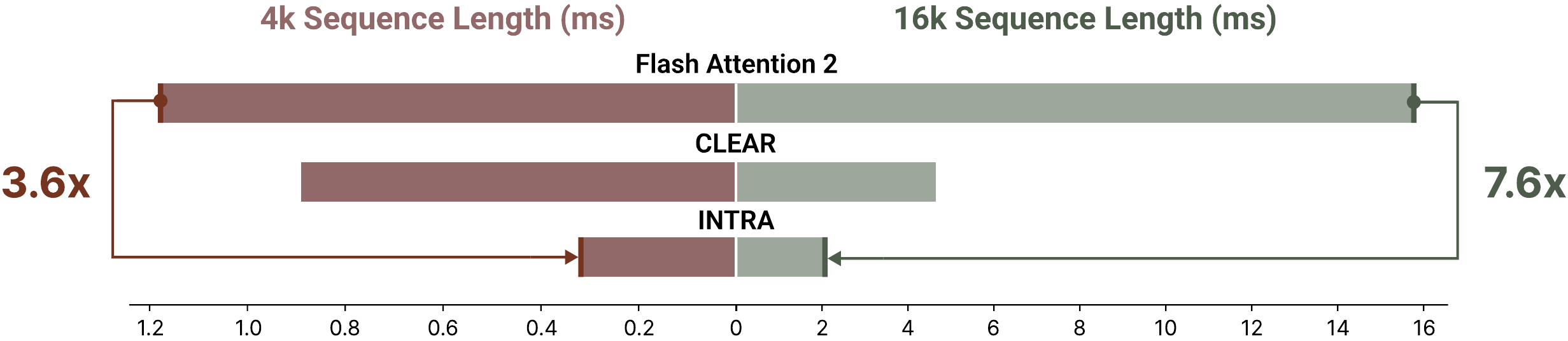}
    \caption{The figure reports the relative speedup achieved by INTRA compared to CLEAR and FlashAttention 2 on 4K and 16K input lengths, evaluated on an A100 GPU. Note that the axis density differs between the left and right sides of the figure to accommodate varying inference times.}
    \label{fig:barchart}
\end{figure}

INTRA combines two CQS-compatible patterns. Under equal-size partitions, Scatter groups same-position tokens across neighborhoods, giving each group one representative per neighborhood. Gather reconnects these channels. If every Scatter group intersects every Gather group, two layers permit graph-theoretic reachability between any source and target. This motivates complementary patterns for efficient sparse communication.

Our contribution is the joint study of structured static attention, GPU execution, and pretrained-model adaptation. Building on factorized and local--global attention~\citep{child2019sparse,tu2022maxvit}, we (1) formulate coverage and overlap objectives within the Intra Sparse Pattern Design (ISPD) Principle and its CQS constraints; (2) implement selected Scatter/Gather patterns using runtime indexing, dense shared-memory packing, and scatter-back; and (3) evaluate their quality--latency trade-offs after adaptation.

INTRA delivers $1.53\times$ faster 2K \texttt{FLUX.1-dev} generation after LoRA self-distillation with better measured image metrics than evaluated CLEAR configurations. Wan2.2-TI2V-5B generation is $1.43\times$ faster end to end at $704\times1280$ with 121 frames (Table~\ref{tab:wan_timing}). On LLaMA-3.1-8B, INTRA accelerates SCROLLS~\citep{shaham2022scrolls} prefill relative to MInference~\citep{jiang2024minference} and XAttention~\citep{xu2025xattention}. Separate 32-layer attention timings at 4K--64K show $3.6$--$8.4\times$ speedups over MoBA and $7.3$--$10.4\times$ over density-constrained NSA (Table~\ref{tab:niah_attention_timing}).

\section{Related Works}

\subsection{Static Sparse Attention}

Static sparse attention uses fixed, input-independent patterns. Sparse Transformer~\citep{child2019sparse} combines local and strided factorizations with short communication paths and interleaves patterns across layers. MaxViT~\citep{tu2022maxvit} combines block and grid attention for local and distributed global interactions. These are close precedents for Scatter/Gather. Sliding-window methods~\citep{jiang2023mistral7b,hassani2023natten,liu2024clear}, Longformer, and BigBird~\citep{beltagy2020longformer,zaheer2020bigbird} provide other local/global constructions. INTRA focuses on CQS-compatible indexing, dense-tile execution, and adaptation of complementary patterns in pretrained models.

\subsection{Dynamic Sparse Attention}

Dynamic sparse methods adapt connectivity to the input or model state. MInference~\citep{jiang2024minference}, FlexPrefill~\citep{lai2025flexprefill}, and XAttention~\citep{xu2025xattention} select heads, blocks, or patterns at runtime. NSA, MoBA, and H2O~\citep{nsa,moba,h2o} also use input-dependent selection or routing, with different training requirements. INTRA evaluates adapted static patterns that avoid runtime selection and target simpler GPU kernels.

\subsection{Other Image Acceleration Methods}

Token pruning and merging methods reduce the effective sequence length for vision transformers by removing or combining tokens~\citep{rao2021dynamicvitefficientvisiontransformers,spvit,tome,toma}. These methods are orthogonal to INTRA, which keeps the token set but changes the sparse communication pattern and kernel implementation.

\subsection{Sparse Video Generation}

Sparse video methods reduce spatiotemporal attention through hardware-aware regions, shifted windows, or spatial/temporal head specialization~\citep{zhang2025sta,liu2022video,sparsevideo}. INTRA formulates a CQS-compatible communication principle applicable across architectures. Our image and language workloads cover high-resolution generation and long-context prefill; the 16K-token setting is also comparable to smaller video models, such as Wan 1.3B at 480p~\citep{wan}. We additionally evaluate video transfer on Wan2.2-TI2V-5B (Appendix~\ref{appendix:wan_vbench}).

\section{Hardware-Efficient Sparse Communication}

\subsection{Computational Query Sets and ISPD Constraints}
We study a class of sparse attention patterns compatible with FlashAttention-style execution. A \textbf{Computational Query Set} (CQS) is a group of query tokens evaluated together, with structured local-to-global mappings for selecting the corresponding key/value tokens. A CQS may use query-dependent key indices, as long as they are generated by structured mappings that can be evaluated uniformly for the group. We call a sparse pattern \textbf{ISPD-compatible} if it satisfies:
\begin{enumerate}
    \item \textbf{Coverage:} The union of all CQSs covers the query-token set. A query token may belong to more than one CQS.
    \item \textbf{Index Mapping:} There exist structured functions $f_q$ and $f_k$ such that, for a given CQS index $i^{\text{cqs}}$,
    \begin{equation}
        i^q_g = f_q(i^q_l, i^{\text{cqs}}), \quad
        i^k_g = f_k(i^k_l, i^q_l, i^{\text{cqs}}).
    \end{equation}
    \item \textbf{Block Alignment:} Each CQS contains a number of queries approximately divisible by the query block size $B$; we use $B=64$ for FlashAttention-style kernels.
\end{enumerate}

\noindent
The ``approximately'' qualifier only accounts for a final partial block when the sequence length is not divisible by $B$. Within this design class, intra-CQS computation avoids cross-thread-group synchronization: queries from the same CQS can be grouped, and the required non-contiguous K/V tokens can be packed contiguously for FlashAttention-style computation. Our Scatter and Gather patterns satisfy these constraints; their explicit $f_q$ and $f_k$ mappings are in Appendix~\ref{appendix:index_function}.

\subsection{Sparse Attention as Communication}
We model a sparse attention layer as a directed information-flow graph over tokens:
\[
G^{(l)}=(V,E^{(l)}), \qquad (j,i)\in E^{(l)} \Longleftrightarrow \text{query token } i \text{ attends to key token } j.
\]
A path follows the direction in which token information can be used by later queries. A \emph{partition-based} sparse layer divides tokens into groups and computes attention within each group. We use this abstraction to reason about Scatter/Gather communication paths.
We assume tokens are partitioned into local neighborhoods
\[
\mathcal{P}=\{P_1,\ldots,P_T\}, \qquad P_t\cap P_{t'}=\varnothing \; (t\neq t'), \qquad \bigcup_{t=1}^T P_t=V.
\]
Here $P_t$ may be an image patch, spatial neighborhood, or short text segment. Our modeling premise is that local attention already provides useful intra-neighborhood mixing; the sparse-design problem is communication between neighborhoods while staying inside the ISPD-compatible design class. We call the sparse groups used to move information between neighborhoods \emph{relay groups}.

\section{Scatter-Gather Relay Design}
\label{sec:scatter-gather}

\subsection{Static Sparse Attention Under Unknown Long-Range Demand}

Dynamic sparse attention can select important keys after observing the input or attention
scores. This adaptivity incurs runtime pattern-selection costs and can introduce irregular
memory access. INTRA instead fixes connectivity in advance, enabling simple local-to-global
index mappings and structured token loading. A static pattern cannot know which long-range token pairs matter for a given input,
head, or layer. It should therefore allocate communication opportunities across possible
long-range interactions, rather than specialize to a single assumed attention pattern.
We express this goal through a worst-case neighborhood-coverage objective.

\subsection{From Direct Edges to Relay Paths}

A sparse layer with per-query budget $g$ can preserve at most $ng$ directed token-token
interactions, while dense attention contains $n^2$ possible interactions. When $g\ll n$,
a static sparse mask cannot directly preserve all potentially important dense edges.

Two sparse layers can provide a mediated path for a missing direct interaction through a relay token:
\[
j \rightarrow r \rightarrow i .
\]
Our analysis counts the relay opportunities connecting source and target neighborhoods.

This observation guides INTRA: the sparse pattern should provide broad and balanced
relay opportunities while remaining compatible with efficient CQS execution.

\subsection{Locality as a Design Motivation}

Nearby image patches often share visual structure, while short text segments provide
coherent local context. This motivates separating two roles: local attention mixes nearby
tokens, and non-local groups provide paths between neighborhoods. When an interaction
involves distant neighborhoods, their local representatives are plausible relay carriers.
The combinatorial objective formalizes this intuition by balancing coverage across
neighborhood pairs.

\subsection{Why Scatter Is Uniform and Non-Contiguous}

Let $N=TR$ tokens be divided into $T\ge2$ local neighborhoods, with integer $R\ge1$:
\[
P_1,\ldots,P_T,
\qquad |P_t|=R.
\]
A group of $T$ tokens covers all $T$ neighborhoods only by taking one representative
from each; repeated samples from one neighborhood leave another uncovered.

Scatter avoids this failure mode. Writing tokens as $(t,r)$, where $t$ is the
neighborhood index and $r$ is the local index, Scatter forms
\[
S_r=\{(t,r):t=1,\ldots,T\},
\qquad r=1,\ldots,R.
\]
Each Scatter group contains one token from every local neighborhood. Thus, every group is
a global low-resolution view of the sequence or image.

Sampling distant neighborhoods makes Scatter non-contiguous. The $R$ groups of size
$T$ compute $RT^2=N^2/R$ attention pairs, retaining a $1/R$ fraction of dense pairs.
At fixed $R$, the complexity is $O(N^2)$. Gather~0, the image variant using $16\times16$ local
neighborhoods, instead uses $O(Nw)$ pairs for a fixed window of $w$ keys per query.

\subsection{Why Gather Is Needed}

Scatter provides global relay groups but separates communication by local index. Tokens
in $S_r$ can exchange information across neighborhoods with other tokens in $S_r$, while
different channels $S_r$ and $S_{r'}$ remain disconnected. This matters because useful
source-side and target-side representatives need not share the same local index: broad
coverage within a channel alone cannot connect every source--target pair.

Gather is introduced to reconnect these Scatter channels. Let
\[
\mathcal G=\{G_b\}_{b=1}^R
\]
be the second partition, and define the Scatter-to-Gather overlap
\[
X_{rb}=|S_r\cap G_b|.
\]
If $X_{rb}=0$, then Scatter channel $S_r$ has no relay witness into Gather group $G_b$.
A good Gather pattern should therefore make these overlaps as balanced as possible,
avoiding weak or missing channel connections.

A simple balanced construction is cyclic:
\[
G_b=\{(t,r): b\equiv t+r \pmod R\}.
\]
Its minimum overlap is $\lfloor T/R\rfloor$. Thus every Scatter group intersects every
Gather group, enabling universal two-layer reachability in this abstract construction,
only when $T\ge R$.

\subsection{Design Summary}

Within equal-size partitions, uniform Scatter maximizes minimum neighborhood-pair
coverage, and balanced Gather maximizes minimum inter-partition overlap. These objectives
motivate complementary patterns under a static communication budget.

Thus INTRA implements the following communication structure:
\[
\text{local mixing}
\;\rightarrow\;
\text{uniform non-contiguous relay}
\;\rightarrow\;
\text{balanced relay reconnection}.
\]
Appendix~\ref{app:relay-theory} proves the coverage and overlap results for the
partition model. We next describe its image and language instantiations.

\subsection{Mapping Back to INTRA}
\label{sec:implemented_patterns}
\begin{figure}[h]
  \centering
  \includegraphics[width=0.78\textwidth]{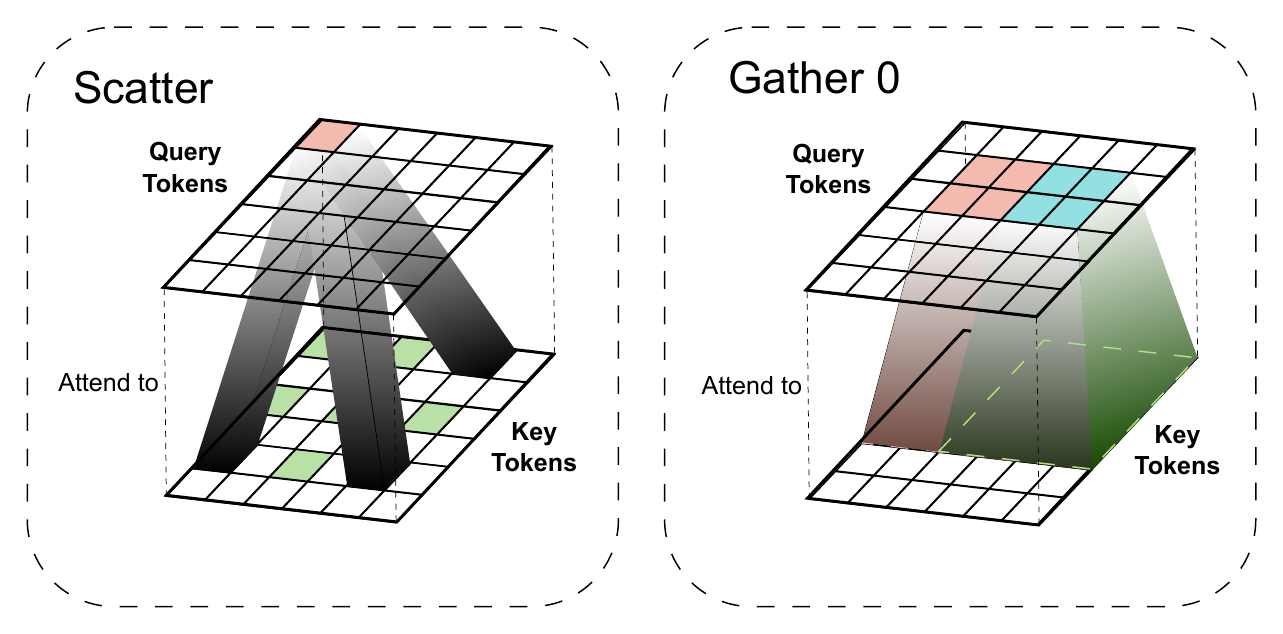}
  \caption{The Scatter and Gather~0 attention patterns used for image generation. Query and key tokens are arranged in a 2D feature map. Scatter groups tokens at the same local position across neighborhoods, while Gather~0 uses a $16\times16$ neighborhood spanning multiple Scatter groups.}
  \label{fig:three_masks}
\end{figure}

For \texttt{FLUX.1-dev}, INTRA interleaves Scatter and Gather~0 (Figure~\ref{fig:three_masks}): Scatter groups same-position tokens across local neighborhoods, and Gather~0 reconnects groups through $16\times16$ neighborhoods. Each layer remains structured while successive patterns provide complementary relay paths. For LLaMA-3.1-8B retrieval, 4 of 32 layers use block-level sparse attention, one every eight layers. These layers select every second contiguous 256-token block, subject to causal masking. The remaining layers use sliding-window attention with a span of 575 tokens and 64 sink tokens~\citep{peng2024streamingllm}. Sparse attention is applied during prefill, followed by dense attention during autoregressive decoding.

For image generation, interleaved Scatter/Gather provides mediated global communication with ISPD-compatible layers. The next section describes the kernel's non-contiguous token loading and contiguous packing.

\section{CQS-Aware Token-Sparse FlashAttention Kernel}
For token-level Scatter/Gather patterns, INTRA implements a CQS-aware variant of FlashAttention~\citep{dao2023flashattention2}. Each CQS is scheduled in query blocks of 64 tokens. Structured mappings $f_q$ and $f_k$ compute global positions at runtime; the kernel loads the selected Q, K, and V tokens, packs them into dense shared-memory tiles, and applies the FlashAttention-style online-softmax core. Outputs are scattered back using the query mapping. This avoids computing and masking unused scores in mixed blocks. Masking within the packed tiles supports full and causal attention.

Scatter and Gather~0 use strided cross-neighborhood access and two-dimensional local offsets, respectively. Gather~1 is the repeated-segment token grouping illustrated in Appendix~\ref{appendix:llm_scatter_gather}; all three share the kernel with different index mappings and underlie the token-level benchmarks. Language retrieval uses the block-level extension in Section~\ref{sec:implemented_patterns}, with selected 256-token blocks in global layers and sliding windows with sinks elsewhere. Appendix~\ref{appendix:overhead} profiles the token-level kernels' indexing, memory traffic, and computation.

\section{Evaluation}
\label{section:evaluation}
We evaluate whether the CQS-compatible relay design is practical in both the \texttt{FLUX.1-dev}~\citep{flux2024} image generation model and the LLaMA 3~\citep{touvron2024llama3} language model. INTRA changes how the model exchanges information between tokens, requiring adaptation; LoRA~\citep{hu2022lora} suffices in our evaluated settings.

\subsection{Experiments}
\label{section:exps}
\textbf{Baseline} \;\; INTRA provides broad communication through fixed sparse patterns. For image generation, we compare with CLEAR, an adapted static sparse-attention method, to evaluate pattern design and execution in the same pretrained-model setting.

For text generation, we compare with MInference~\citep{jiang2024minference} and XAttn~\citep{xu2025xattention} on the long-context SCROLLS benchmark~\citep{shaham2022scrolls}.

\textbf{Experiment Setup} \;\; We fine-tune \texttt{FLUX.1-dev} using CLEAR's 10K self-generated image-text dataset. Following CLEAR, we sample 5k COCO~\citep{coco} validation prompts and evaluate generated images using FID~\citep{heusel2017fid,fid}, LPIPS~\citep{lpips}, and CLIPI~\citep{clip}. For LLaMA 3, we fine-tune and evaluate on SCROLLS~\citep{shaham2022scrolls}, covering GovReport~\citep{gov_report}, Qasper~\citep{qasper}, QMSum~\citep{qmsum}, QuALITY~\citep{quality}, SummScreen-FD~\citep{summscreenfd}, and ContractNLI~\citep{contractnli}. These tasks fit our 16K-token fine-tuning budget and provide reliable validation sets.

\textbf{Resource Usage and Practicality} \;\; All fine-tuning and evaluation use NVIDIA A100 GPUs. \texttt{FLUX.1-dev} fine-tuning takes 43 GPU hours. The matched language retrieval protocol uses rank-16 LoRA and takes about one hour on eight GPUs per run (approximately 8 GPU hours). Appendix~\ref{appendix:retrieval_adaptation} gives the adaptation recipe and distinguishes it from the earlier SCROLLS fine-tuning setup.

\textbf{Results} \;\; Table~\ref{tab:speed_compare_table_flux} reports image attention-kernel speedups; token-level language timings are in Table~\ref{tab:latency_scroll_benchmark}(a) and Appendix~\ref{appendix:overhead}. At 2K, INTRA achieves the best measured image metrics and lowest generation latency among evaluated sparse configurations: 43.72 seconds and FID 15.67 versus CLEAR ($r=8$)'s 32.46 (Table~\ref{tab:image_benchmark}). Table~\ref{tab:latency_scroll_benchmark} reports single attention operation timings alongside separate SCROLLS quality results: INTRA has the lowest latency at all three context lengths and the highest QuALITY score. Appendix~\ref{appendix:speed_quality} gives group-count sweeps and the summary calculations.

For video transfer, INTRA reduces Wan2.2-TI2V-5B~\citep{wan,wan22} generation time from 340.9 to 238.4 seconds ($1.43\times$) at $704\times1280$ with 121 frames on A100-SXM4-80GB. At the VBench~\citep{vbench} setting ($480\times832$, 49 frames), latency falls from 44.1 to 41.2 seconds; evaluation on 944 prompts with five samples each yields a 13-dimension mean of 0.4677 versus dense attention's 0.4745. Table~\ref{tab:wan_timing} combines end-to-end, forward, and attention timings; Appendix~\ref{appendix:wan_vbench} gives per-dimension quality results.

\begin{table*}[t!]
\centering
\caption{Single full-attention latency and speedup on A100. INTRA interleaves Scatter/Gather for image generation.}
\vspace{0.3em}
\renewcommand{\arraystretch}{1.2}
\resizebox{\textwidth}{!}{
\begin{tabular}{lcccccccc}
\toprule
\textbf{Methods} &  \textbf{Seq Len}  &  \textbf{Config} & \textbf{GFLOPS} & \textbf{Latency(ms)} & \textbf{Speedup} & \textbf{FID $\downarrow$} \\
\midrule
FlashAttention 2 & \multirow{3}{*}{4K} & - & 260.9 & 1.181 & - & - \\

CLEAR &  & r=8 & 63.5 & 0.893 & $1.32\times$ & 11.70\\
INTRA  &  & Interleave Scatter/Gather & 50.75 & 0.324 & $3.64\times$ & 11.74 \\

\midrule
\midrule

FlashAttention 2 & \multirow{3}{*}{16K} & - & 3507.9 & 15.764 & - & - \\

CLEAR &  & r=8 & 246.2 & 4.625 & $3.41\times$ & 32.46\\
INTRA  &  & Interleave Scatter/Gather & 345.5 & 2.056 & $7.67\times$ & 15.67 \\
\bottomrule
\end{tabular}}
\label{tab:speed_compare_table_flux}
\end{table*}

\begin{table*}[t!]
\centering
\caption{End-to-end latency and image quality at 1K/2K. CLEAR (r=4--8) uses 1/4-resolution K/V and radius 8; CLEAR (r=48) is slower than dense \texttt{FLUX.1-dev}.}
\vspace{0.2em}
\small  %
\renewcommand{\arraystretch}{1.2}
\resizebox{1.0\textwidth}{!}{  %
\begin{tabular}{lcccc|cccc}
\toprule
& \multicolumn{4}{c|}{\textbf{1024$\times$1024 Image Generation}} & \multicolumn{4}{c}{\textbf{2048$\times$2048 Image Generation}} \\
\cmidrule(lr){2-5} \cmidrule(lr){6-9}
  \textbf{Method} & \textbf{FID ($\downarrow$)} & \textbf{LPIPS ($\downarrow$)} & \textbf{CLIPI ($\uparrow$)} & \textbf{Gen Time(s)} & \textbf{FID ($\downarrow$)} & \textbf{LPIPS ($\downarrow$)} & \textbf{CLIPI ($\uparrow$)} & \textbf{Gen Time(s)}\\
\midrule
Original Model & - & - & - & 13.42 & - & - & - & 66.78\\
\midrule
CLEAR (r=8) & \textbf{11.70} & 49.96 & 0.8995 & 12.85 & 32.46 & 61.62 & 0.8046 & 47.81 \\
CLEAR (r=4-8) & - & - & - & 13.73 & 19.27 & 54.60 & 0.8467 & 52.07 \\
INTRA       & 11.74 & \textbf{49.46} & \textbf{0.9045} & \textbf{11.97} & \textbf{15.67} & \textbf{51.99} & \textbf{0.8559} & \textbf{43.72} \\
\bottomrule
\end{tabular}}
\label{tab:image_benchmark}
\end{table*}

Appendix~\ref{appendix:image_ablations} compares with a smaller dense diffusion model; Appendix~\ref{appendix:visual_compare} provides visual comparisons.

\begin{table}[!htb]
\centering
\caption{Attention operation speed and separate SCROLLS quality. INTRA achieves \textbf{2.23$\times$ the speed of XAttention} and \textbf{14.94$\times$ the speed of MInference} (geometric means, 8K--32K). Respective unweighted mean-score gaps are 0.87 and 0.42 points.}
\label{tab:latency_scroll_benchmark}
\small
\setlength{\tabcolsep}{3pt}
\renewcommand{\arraystretch}{1.0}
\begin{subtable}{\textwidth}
\centering
\caption{Single causal-attention operation latency (ms$\,\downarrow$) on A100.  INTRA uses token-level Scatter/Gather.}
\begin{tabular*}{\textwidth}{@{\extracolsep{\fill}}lrrr@{}}
\toprule
\textbf{Method} & \textbf{8K} & \textbf{16K} & \textbf{32K} \\
\midrule
XAttention ($S=16$) & 1.91 & 4.22 & 8.03 \\
MInference & 14.30 & 26.42 & 51.20 \\
\textbf{INTRA} & \textbf{0.51} & \textbf{1.75} & \textbf{6.50} \\
\bottomrule
\end{tabular*}
\end{subtable}
\par\smallskip
\begin{subtable}{\textwidth}
\centering
\caption{SCROLLS task scores and unweighted six-task mean ($\uparrow$).}
\begin{tabular*}{\textwidth}{@{\extracolsep{\fill}}lrrrrrrr@{}}
\toprule
\textbf{Method} & \textbf{GovReport} & \textbf{Qasper} & \textbf{QMSum} & \textbf{QuALITY} & \shortstack{\textbf{SummScreen}\\\textbf{-FD}} & \shortstack{\textbf{Contract}\\\textbf{NLI}} & \shortstack{\textbf{Unweighted}\\\textbf{mean}} \\
\midrule
XAttention & 31.16 & 39.94 & 24.93 & 27.5 & 20.27 & \textbf{77.5} & \textbf{36.88} \\
MInference & \textbf{31.44} & \textbf{39.99} & \textbf{25.18} & 24.5 & \textbf{20.40} & 77.1 & 36.44 \\
\textbf{INTRA} & 27.77 & 39.64 & 24.45 & \textbf{29.0} & 19.71 & 75.5 & 36.01 \\
\bottomrule
\end{tabular*}
\end{subtable}
\end{table}

\textbf{Needle Retrieval and Attention Efficiency.} \;\; We evaluate LLaMA-3.1-8B on needle-in-a-haystack (NIAH)~\citep{kamradt2023niah} retrieval at nine depths and 4K--32K contexts. Dense attention, MoBA~\citep{moba}, NSA~\citep{nsa}, and INTRA share the LoRA budget and teacher-forced exact-match protocol; each model reuses one adapter trained on disjoint data. INTRA scores 0.923 on average and 0.918 at 32K. Table~\ref{tab:niah_retrieval} presents retrieval scores alongside separate attention-only timings: across 4K--64K, INTRA is $3.6$--$8.4\times$ faster than MoBA and $7.3$--$10.4\times$ faster than tuned NSA.

\begin{table}[!htb]
\centering
\caption{LLaMA-3.1-8B retrieval quality and separate attention timing experiments.}
\label{tab:niah_retrieval}
\small
\renewcommand{\arraystretch}{1.0}
\begin{subtable}{\textwidth}
\centering
\caption{Mean NIAH with matched LoRA adaptation (9 depths; 4K--32K).}
\begin{tabular}{lcccc}
\toprule
\textbf{Method} & Dense + LoRA & MoBA & NSA & INTRA \\
\midrule
\textbf{Mean NIAH $\uparrow$} & 1.000 & 0.194 & 0.325 & \textbf{0.923} \\
\bottomrule
\end{tabular}
\end{subtable}
\par\smallskip
\begin{subtable}{\textwidth}
\centering
\caption{Attention latency (ms$\,\downarrow$) on A100-SXM4-80GB.}
\label{tab:niah_attention_timing}
\begin{tabular}{lrrrr}
\toprule
\textbf{Context} & \shortstack{\textbf{Dense}\\(SDPA)} & \shortstack{\textbf{INTRA}\\(\texttt{g2s4})} & \textbf{MoBA} & \shortstack{\textbf{NSA (tuned)}\\$\rho\leq0.125$} \\
\midrule
4K & 27.4 & \textbf{11.3} & 95.2 & 117.7 \\
16K & 381.0 & \textbf{59.7} & 349.7 & 569.5 \\
64K & 6043.3 & \textbf{577.2} & 2073.0 & 4235.6 \\
\bottomrule
\end{tabular}
\end{subtable}
\par\smallskip
\begin{minipage}{\textwidth}
\footnotesize
Attention only: sum of 32 per-layer medians (50 repetitions), bfloat16, batch 1. INTRA \texttt{g2s4} uses four Scatter $G=2$ layers and 28 width-576 windows; NSA is tuned under density cap $\rho\leq0.125$. Retrieval and timing configurations are specified in Section~\ref{sec:implemented_patterns} and Appendix~\ref{appendix:moba_speed}, respectively.
\end{minipage}
\end{table}

\textbf{Cross-Model Evaluation.} \;\; On Qwen3-8B~\citep{qwen3}, INTRA achieves a mean NIAH score of 0.912 versus dense attention's 1.000, using the same evaluation protocol and LoRA recipe and budget. Qwen3-8B uses five block-level sparse layers among 36, with sparse-layer retained density tuned for each model family.

\subsection{Ablation Study}

\textbf{Global-Layer Ablation.} \;\; We fix LLaMA-3.1-8B, training data, LoRA rank and steps, and four layer positions; the other 28 layers retain the same windows and sink tokens. Only attention at the selected positions changes. Sliding-window, INTRA block-sparse, and dense fillings obtain mean NIAH scores of 0.028, 0.923, and 1.000 at overall retained densities of 0.069, 0.123, and 0.185 (Table~\ref{tab:global_layer_ablation}). INTRA thus retains 92.3\% of the dense-filled control's retrieval score with about two-thirds of its retained attention entries. This measures the quality--density trade-off with fixed layer placement and adaptation.

\begin{table}[htbp]
\centering
\caption{LLaMA-3.1-8B attention-pattern ablation at four fixed layers with matched training budgets. The other 28 layers use windows with sinks; density covers all 32 layers.}
\label{tab:global_layer_ablation}
\small
\renewcommand{\arraystretch}{0.95}
\begin{tabular}{lcc}
\toprule
\textbf{Pattern in selected layers} & \textbf{Overall retained density} & \textbf{Mean NIAH $\uparrow$} \\
\midrule
Sliding window & 0.069 & 0.028 \\
INTRA block-level sparse attention & 0.123 & 0.923 \\
Dense attention & 0.185 & 1.000 \\
\bottomrule
\end{tabular}
\end{table}

\textbf{Local Windows and Global Reach.} \;\; To test whether increasing local capacity suffices, Appendix~\ref{appendix:window_width} varies window width from 575 to 4096 in all 32 layers. At 32K, width 4096 scores 0.111 despite reaching 1.000 at 4K; INTRA's four global sparse layers retain 0.918 at 32K. Together with the configuration comparison in Appendix~\ref{appendix:language_configuration}, this supports allocating attention to global reach rather than only widening local windows.

\textbf{Image Pattern and Sparsity.} \;\; Table~\ref{tab:ablation-combined} separately tests using either Scatter or Gather alone and varying the Scatter group count during interleaving. Scatter alone yields FID 213.47, Gather~0 improves this to 13.23, and interleaving reaches 11.75, with consistent LPIPS and CLIPI improvements. This contrast is consistent with the complementary roles in Section~\ref{sec:scatter-gather}: Scatter supplies cross-neighborhood communication, while Gather reconnects otherwise isolated Scatter channels.

With interleaving retained, increasing Scatter groups from 4 to 8 to 16 reduces attended tokens but raises FID from 10.99 to 11.75 to 12.85; eight groups is an intermediate quality--sparsity choice. Switching every two layers gives the best measured metrics among intervals of one to four layers (Appendix~\ref{appendix:image_ablations}). Appendix~\ref{appendix:additional_results} collects the supplementary experiments and adaptation studies.

\begin{table}[H]
\caption{Ablation study: pattern choice and number of Scatter groups during interleaving.}
\vspace{0.2em}
\scriptsize
\setlength{\tabcolsep}{2pt}
\renewcommand{\arraystretch}{1.2}
\makebox[\textwidth][c]{%
  \begin{subtable}[t]{0.43\textwidth}
  \centering
  \caption{Use Scatter/Gather or both}
  \begin{tabular}{lccc}
  \toprule
   & \textbf{FID ($\downarrow$)} & \textbf{LPIPS ($\downarrow$)} & \textbf{CLIPI ($\uparrow$)}\\
  \midrule
  Scatter & 213.47 & 64.91 & 0.67\\
  Gather 0 & 13.23 & 54.10 & 0.88\\
  Scatter + Gather 0 & \textbf{11.75} & \textbf{49.46} & \textbf{0.90}\\
  \bottomrule
  \end{tabular}
  \end{subtable}%
  \hspace{0.04\textwidth}
  \begin{subtable}[t]{0.43\textwidth}
  \centering
  \caption{Number of groups in Scatter}
  \begin{tabular}{lccc}
  \toprule
   & \textbf{FID ($\downarrow$)} & \textbf{LPIPS ($\downarrow$)} & \textbf{CLIPI ($\uparrow$)}\\
  \midrule
  4 groups & \textbf{10.99} & \textbf{48.29} & \textbf{0.908}\\
  8 groups & 11.75 & 49.46 & 0.905\\
  16 groups & 12.85 & 49.88 & 0.896\\
  \bottomrule
  \end{tabular}
  \end{subtable}%
}
\label{tab:ablation-combined}
\end{table}

\subsection{Limitations}
\label{section:limitation}
Our partition analysis optimizes coverage and overlap, not attention approximation or downstream quality. The deployed geometric and causal masks impose additional constraints; causal connectivity is restricted to preceding tokens. Comparisons with other factorized patterns and larger models remain future work.

Quality is task-dependent: INTRA trails both SCROLLS baselines on five of six tasks. Relative to adapted dense attention, GovReport falls from 31.05 to 27.77, while Qasper changes from 40.09 to 39.64 (Table~\ref{tab:llm_no_finetune}). Image comparisons do not establish dense-quality parity. Wan's aesthetic and imaging scores decrease by 0.0194 and 0.0564; shared zeros remain unexplained. Video quality is evaluated only at 49 frames.

Retrieval measures teacher-forced exact match after shared-budget adaptation, rather than free-generation or zero-shot performance or general rankings of MoBA/NSA. Layer ablations compare quality and density, not matched FLOPs; configuration sweeps vary both density and layer allocation. The SD~1.5 comparison also differs in architecture and training data. Quality scores lack confidence intervals; timing repetitions do not measure quality uncertainty.

Runtime indexing and non-contiguous loads add overhead; unchanged components limit end-to-end gains (Appendix~\ref{appendix:overhead}). Token kernels and block-sparse retrieval use distinct configurations. MoBA/NSA timing comparisons cover attention only, not full-model prefill. Reshaping changes scheduling and occupancy, preventing isolated overhead measurement.

\section{Conclusion}
We introduced INTRA as a sparse communication design under ISPD/CQS constraints. In a partition-based relay model, Scatter gives max-min local-neighborhood coverage and Gather gives balanced two-layer reconnection. Together, they provide a communication criterion for complementary static patterns.

The CQS-aware kernel turns these patterns into dense-tile GPU computation through runtime indexing and non-contiguous token packing. After adaptation, INTRA delivers $1.53\times$ faster 2K \texttt{FLUX.1-dev} generation and better measured image metrics than evaluated CLEAR configurations. Language experiments demonstrate faster prefill and effective long-range retrieval with a small allocation of global sparse layers, supporting structured static attention as a practical acceleration strategy.

\clearpage
\subsection*{AI use statement}
Generative AI assistants were used for polishing and reference checking. Our adaptation protocols also use model-generated data and dense-teacher outputs, as described in Section~\ref{section:exps} and Appendix~\ref{appendix:retrieval_adaptation}. AI-assisted revisions were checked against cited primary sources and reported tables where applicable; these checks do not constitute independent reruns of the experiments. The authors proposed and determined the research questions, central ideas, methodology, and final technical claims. The authors take responsibility for the final text, claims, citations, and artifacts, including material prepared with AI assistance.

\bibliographystyle{iclr2027_conference}
\bibliography{references}

\newpage
  \appendix

\section{Token-Level Scatter and Gather 1 Patterns}
\label{appendix:llm_scatter_gather}

\begin{figure}[h]
  \centering
  \includegraphics[width=1\textwidth, trim=80 70 80 80, clip]{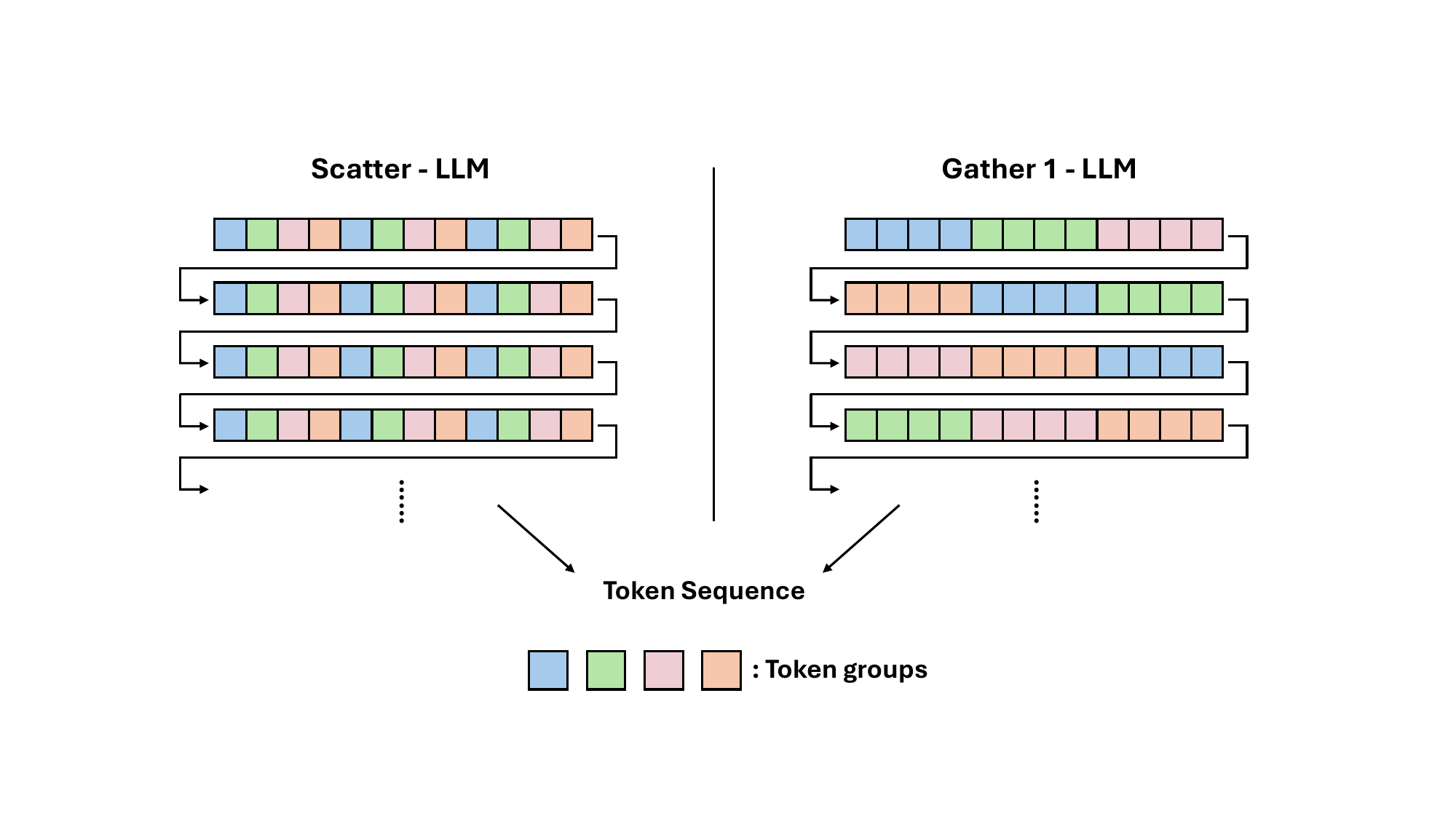}
  \caption{Token-level Scatter and Gather 1 patterns with a grouping factor of 4. Tokens are partitioned into 4 groups, shown by color, and both patterns compute attention within each group. These illustrate kernel variants rather than the block-level mask used in the retrieval experiments.}
  \label{fig:llm_mask}
\end{figure}

Figure~\ref{fig:llm_mask} illustrates the token-level Scatter and Gather 1 variants used in the kernel benchmarks. The language retrieval experiments instead use the block-level global layers and sliding-window layers with sink tokens described in Section~\ref{sec:implemented_patterns}.

\section{GPU Memory Hierarchy}
\label{appendix:gpu_mem_hierarchy}
NVIDIA GPUs use a memory hierarchy consisting primarily of global memory, shared memory, and registers. Global memory is large but high-latency; shared memory is faster and shared within a thread block; registers are fastest and private to individual threads. INTRA changes which global-memory tokens are loaded, then packs the selected tokens contiguously in shared memory before applying FlashAttention-style computation.

\section{Sparse Pattern Index Function}
\label{appendix:index_function}
\subsection{Scatter}
\label{appendix:scatter}
For the LLM Scatter pattern with query block size 64 and 8 groups, let $i^{\text{cqs}}$ be the CQS index. The local-to-global mappings are
\begin{equation}
f_q(i^{q}_{l}, i^{\text{cqs}}) =  i^{\text{cqs}} + 8 \cdot i^{q}_{l},
\end{equation}
\begin{equation}
f_k(i^{k}_{l}, i^{q}_{l}, i^{\text{cqs}}) = i^{\text{cqs}} + 8 \cdot i^{k}_{l}.
\end{equation}

\subsection{Gather 0}
\label{appendix:gather0}
Assume an image feature map with width $W$ and height $H$, both at least 16 and divisible by 8. Each CQS is an $8\times8$ query patch attending to a $16\times16$ image-key window. We use zero-based, row-major indexing within the image. For CQS index $i^{\text{cqs}}$, let $r=\lfloor i^{\text{cqs}}/(W/8)\rfloor$ and $c=i^{\text{cqs}}\bmod(W/8)$. For $0\leq i_l^q<64$,
\begin{equation}
f_q(i^{q}_{l}, i^{\text{cqs}})
=\left(8r+\left\lfloor i_l^q/8\right\rfloor\right)W+8c+(i_l^q\bmod8).
\end{equation}
The window is shifted inward at image boundaries. Its top-left row and column are
\begin{equation}
r_0=\min\!\bigl(\max(8r-4,0),H-16\bigr),\qquad
c_0=\min\!\bigl(\max(8c-4,0),W-16\bigr).
\end{equation}
All queries in the CQS share the same image-key mapping, with $0\leq i_l^k<256$:
\begin{equation}
f_k(i_l^k,i_l^q,i^{\text{cqs}})
=\left(r_0+\left\lfloor i_l^k/16\right\rfloor\right)W+c_0+(i_l^k\bmod16).
\end{equation}
The kernel packs the window into four 64-key tiles ($i_l^k=64t+j$, $0\leq t<4$, $0\leq j<64$), each covering four rows of 16 keys. Thus every image index lies in $[0,HW)$, including at corners. In the released FLUX implementation, $H=W\in\{64,128\}$ and 512 text tokens precede the image tokens, so these image indices receive an offset of 512. Every query also attends to all text keys; text queries retain their Scatter image-key groups.

\section{Formal Relay Analysis for Static CQS Design}
\label{app:relay-theory}

This appendix formalizes the relay-design intuition used in Section~\ref{sec:scatter-gather}.
The mask-level analysis establishes uniform Scatter and balanced Gather as minimax
allocations for neighborhood coverage and inter-partition overlap under CQS-style
partition constraints.

\subsection{Setup}

Let $T\ge2$ and $R\ge1$ be integers. Partition $N=TR$ tokens into $T$ local neighborhoods:
\[
P_1,\ldots,P_T,
\qquad |P_t|=R.
\]
We consider static partition-based CQS groups. Scatter consists of $R$ relay groups
\[
\mathcal S=\{S_a\}_{a=1}^R,
\qquad |S_a|=T,
\]
which partition the token set. Define
\[
c_{a,t}=|S_a\cap P_t|.
\]

The proof uses token counts and intersections of fixed sets. The local neighborhoods
specify the partition over which coverage is evaluated.

\subsection{Uniform Scatter as Minimax Neighborhood-Pair Coverage}

For a pair of neighborhoods $(P_s,P_t)$, define the relay opportunity count
\[
K_{s,t}
=
\sum_{a=1}^R c_{a,s}c_{a,t}.
\]
This counts how many ordered source-target representative pairs from $P_s$ and $P_t$ are
co-grouped in some Scatter relay group.

Let $W_{s,t}\ge 0$ be an arbitrary normalized long-range demand matrix over ordered
neighborhood pairs:
\[
\sum_{s\ne t} W_{s,t}=1.
\]
Define the demand-weighted coverage of $\mathcal S$ as
\[
\Phi(\mathcal S;W)
=
\sum_{s\ne t} W_{s,t}K_{s,t}.
\]
Since the static mask does not know $W$ in advance,
\[
\min_W \Phi(\mathcal S;W)
=
\min_{s\ne t}K_{s,t}.
\]
Thus robustness to arbitrary neighborhood-level demand reduces to maximizing the weakest
relay opportunity count.

\paragraph{Proposition \thesection.1.}
For any partition into $R$ relay groups of size $T$,
\[
\min_{s\ne t}K_{s,t}\le R.
\]
Uniform Scatter achieves equality by taking exactly one token from every neighborhood in
every relay group:
\[
|S_a\cap P_t|=1
\qquad
\forall a,t.
\]
Consequently,
\[
\max_{\mathcal S}\min_W \Phi(\mathcal S;W)
=
R.
\]

\paragraph{Proof.}
For each relay group $S_a$,
\[
\sum_{t=1}^T c_{a,t}=T.
\]
Therefore,
\[
\sum_{s\ne t} c_{a,s}c_{a,t}
=
\left(\sum_t c_{a,t}\right)^2-\sum_t c_{a,t}^2
\le
T^2-T,
\]
because $\sum_t c_{a,t}^2\ge T$ for nonnegative integers summing to $T$. Summing over
all $R$ groups gives
\[
\sum_{s\ne t}K_{s,t}
\le
RT(T-1).
\]
There are $T(T-1)$ ordered neighborhood pairs, so the average value of $K_{s,t}$ is at
most $R$. Hence
\[
\min_{s\ne t}K_{s,t}\le R.
\]
Uniform Scatter has $c_{a,t}=1$ for all $a,t$, so
\[
K_{s,t}
=
\sum_{a=1}^R 1\cdot 1
=
R
\]
for every ordered pair $s\ne t$, attaining the upper bound.
\hfill $\square$

\paragraph{Interpretation.}
Any non-uniform relay group that samples one neighborhood multiple times must
under-sample another. This can improve some neighborhood pairs only by weakening others.
If long-range demand is unknown, the weakened pair becomes the worst case. Uniform Scatter
is therefore the no-prior robust static allocation of relay opportunities.

\subsection{Balanced Gather as Minimax Channel Reconnection}

Scatter creates global relay groups but separates them by local index. Write
\[
S_r=\{(t,r):t=1,\ldots,T\},
\qquad r=1,\ldots,R.
\]
Let Gather be another partition
\[
\mathcal G=\{G_b\}_{b=1}^R,
\qquad |G_b|=T,
\]
and define the overlap matrix
\[
X_{rb}=|S_r\cap G_b|.
\]
The entry $X_{rb}$ counts the number of relay witnesses from Scatter channel $S_r$ into
Gather group $G_b$. If $X_{rb}=0$, then that channel-to-group connection is absent.

For arbitrary reconnection demand, the weakest entry controls the worst case. Since each
Scatter group has size $T$,
\[
\sum_{b=1}^R X_{rb}=T.
\]
Therefore,
\[
\min_b X_{rb}
\le
\left\lfloor \frac{T}{R}\right\rfloor.
\]
Thus no Gather partition can make the weakest overlap larger than
$\lfloor T/R\rfloor$.

A cyclic Gather construction achieves this value:
\[
G_b=\{(t,r): b\equiv t+r \pmod R\}.
\]
For fixed $r$ and $b$,
\[
|S_r\cap G_b|
=
|\{t:b\equiv t+r\pmod R\}|
\in
\left\{
\left\lfloor \frac{T}{R}\right\rfloor,
\left\lceil \frac{T}{R}\right\rceil
\right\}.
\]
Thus cyclic Gather maximizes the weakest Scatter-to-Gather overlap under the same
equal-size partition constraint. Universal two-layer connectivity additionally requires
$T\ge R$, so that $\lfloor T/R\rfloor\ge1$. If $T<R$, zero overlaps remain and the
construction does not provide universal two-layer connectivity.

\subsection{Relation to Implemented Patterns}

The results optimize neighborhood-pair coverage and inter-partition overlap within
equal-size partitions. Uniform Scatter distributes representatives across neighborhoods,
and cyclic Gather balances connections between the resulting channels. These objectives
guide the implementations in Section~\ref{sec:implemented_patterns}: image generation
combines Scatter with geometric Gather~0 neighborhoods, while language retrieval combines
causal global block attention with local windows and sink tokens. The global-layer and
image-pattern ablations evaluate these complementary roles after adaptation.

\paragraph{Balanced Gather reconnection.}
Given fixed Scatter groups and fixed Gather group sizes, the row and column sums of $X_{ab}$ are fixed by the partition sizes. Maximizing $\min_{a,b}X_{ab}$ therefore favors balancing the overlap matrix and avoiding zero entries. When every entry is positive, the two-layer reachability result applies.

\section{Overhead Analysis}
\label{appendix:overhead}

\begin{table}[htbp]
\centering
\caption{Supplementary single causal-attention operation latency (ms) on A100. The FlashAttention 2 baseline and INTRA 4K result complement Table~\ref{tab:latency_scroll_benchmark}(a), which reports the 8K--32K sparse-method timings. INTRA uses token-level Scatter/Gather.}
\label{tab:single_att_causal_kernel_speed}
\small
\renewcommand{\arraystretch}{1.1}
\begin{tabular}{lcccc}
\toprule
\textbf{Method} & \textbf{4K} & \textbf{8K} & \textbf{16K} & \textbf{32K} \\
\midrule
FlashAttention 2 & 0.82 & 2.86 & 10.57 & 40.96 \\
INTRA & 0.16 & \multicolumn{3}{c}{See Table~\ref{tab:latency_scroll_benchmark}(a)} \\
\bottomrule
\end{tabular}
\end{table}

\begin{table}[htbp]
\centering
\caption{GFLOPS entries accompanying the single causal-attention timings in Tables~\ref{tab:latency_scroll_benchmark}(a) and~\ref{tab:single_att_causal_kernel_speed}. Dashes indicate unreported settings.}
\label{tab:causal_kernel_gflops}
\small
\begin{tabular}{lcccc}
\toprule
\textbf{Method} & \textbf{4K} & \textbf{8K} & \textbf{16K} & \textbf{32K} \\
\midrule
FlashAttention 2 & 137.4 & 549.8 & 2199.1 & 8796.3 \\
MInference & -- & 329.8 & 1275.4 & 4398.1 \\
XAttn S=16 & -- & 153.6 & 614.2 & 2615.1 \\
INTRA & 17.2 & 68.7 & 274.9 & 1099.5 \\
\bottomrule
\end{tabular}
\end{table}
This appendix profiles the token-level Scatter/Gather kernels. The language-sequence patterns are illustrated in Appendix~\ref{appendix:llm_scatter_gather}. With 8 groups, the ideal attention speedup is $8\times$. Actual Scatter/Gather speedups approach this value as sequence length grows, indicating that fixed overheads become less significant for longer contexts. The block-level mask construction cost for the language retrieval configuration is reported separately in Appendix~\ref{appendix:mask_construction}.

\begin{table}[h]
\centering
\caption{Ideal vs. actual speedups for the token-level Scatter/Gather kernels on A100.}
\vspace{0.5em}
\small
\begin{tabular}{lccc}
\toprule
\textbf{Pattern} & \textbf{Seq. Len.} & \textbf{Ideal} & \textbf{Actual} \\
\midrule
Scatter & 4K  & $8\times$ & $4.81\times$ \\
Gather 1 & 4K  & $8\times$ & $5.84\times$ \\
Scatter & 8K  & $8\times$ & $5.43\times$ \\
Gather 1 & 8K  & $8\times$ & $5.83\times$ \\
Scatter & 16K & $8\times$ & $5.90\times$ \\
Gather 1 & 16K & $8\times$ & $6.15\times$ \\
Scatter & 32K & $8\times$ & $6.22\times$ \\
Gather 1 & 32K & $8\times$ & $6.37\times$ \\
\bottomrule
\end{tabular}
\label{tab:ideal_actual_speedup}
\end{table}

Nsight profiling in Table~\ref{tab:nsight_profile} characterizes the balance between memory traffic and computation after reducing the QK$^\top$ workload. INTRA has higher memory throughput and lower compute throughput than dense FlashAttention, consistent with its reduced arithmetic workload. Gather~0's more contiguous loads produce fewer uncoalesced global-memory transactions than Scatter's strided loads.

\begin{table}[h]
\centering
\caption{Nsight profiling at 16K in the image setting. Throughput numbers are reported by NVIDIA profiling tools at the kernel level.}
\vspace{0.5em}
\small
\begin{tabular}{lccc}
\toprule
\textbf{Method} & \textbf{Memory Throughput} & \textbf{Compute Throughput} & \textbf{Time (ms)} \\
\midrule
FlashAttention & 49.95 & 70.69 & 15.80 \\
INTRA Scatter & 86.31 & 56.64 & 3.04 \\
INTRA Gather 0 & 79.48 & 51.19 & 1.07 \\
\bottomrule
\end{tabular}
\label{tab:nsight_profile}
\end{table}

Table~\ref{tab:equally_sparse} compares INTRA with an equally sparse dense-FlashAttention simulation. The simulation reshapes inputs to match sparsity, providing a computational reference for the Scatter and Gather~1 timings.

\begin{table}[h]
\centering
\caption{INTRA compared with an equally sparse dense-FlashAttention simulation.}
\vspace{0.5em}
\small
\begin{tabular}{llcc}
\toprule
\textbf{Pattern} & \textbf{Seq. Len.} & \textbf{INTRA (ms)} & \textbf{Equally Sparse (ms)} \\
\midrule
Scatter & 4K  & 0.17 & 0.19 \\
Scatter & 8K  & 0.53 & 0.52 \\
Scatter & 16K & 1.79 & 1.65 \\
Scatter & 32K & 6.58 & 5.73 \\
Gather 1 & 4K  & 0.29 & 0.22 \\
Gather 1 & 16K & 1.07 & 0.79 \\
\bottomrule
\end{tabular}
\label{tab:equally_sparse}
\end{table}

Table~\ref{tab:remove_attention_latency} isolates the attention contribution by replacing attention with a dummy operation. The left subtable reports full \texttt{FLUX.1-dev} generation latency at 4K and 16K, while the right subtable reports LLaMA 3 prompt-prefill latency from 4K to 32K. INTRA therefore targets the attention-heavy generation and prefill stages; autoregressive decoding is more memory-bound, and batch parallelism remains orthogonal because the sparse pattern is applied independently within each sequence.

\begin{table}[h]
\centering
\caption{End-to-end latency with and without attention.}
\vspace{0.5em}
\small
\makebox[\textwidth]{%
  \begin{subtable}[t]{0.38\textwidth}
  \centering
  \caption{\texttt{FLUX.1-dev} generation}
  \begin{tabular}{lcc}
  \toprule
  \textbf{Setting} & \textbf{4K} & \textbf{16K} \\
  \midrule
  Original & 13.4s & 66.8s \\
  Remove attention & 11.6s & 41.2s \\
  \bottomrule
  \end{tabular}
  \end{subtable}%
  \hfill
  \begin{subtable}[t]{0.56\textwidth}
  \centering
  \caption{LLaMA 3 prefill}
  \begin{tabular}{lcccc}
  \toprule
  \textbf{Setting} & \textbf{4K} & \textbf{8K} & \textbf{16K} & \textbf{32K} \\
  \midrule
  Original & 0.31s & 0.64s & 1.41s & 3.47s \\
  Remove attention & 0.27s & 0.53s & 1.05s & 2.07s \\
  \bottomrule
  \end{tabular}
  \end{subtable}%
}
\label{tab:remove_attention_latency}
\end{table}

The latency measurements in this overhead analysis are averaged over 2,000 runs. Model-quality scores are reported as point estimates.

\section{Speed--Quality Trade-off}
\label{appendix:speed_quality}
\begin{figure}[htbp]
  \centering
  \includegraphics[width=1\textwidth]{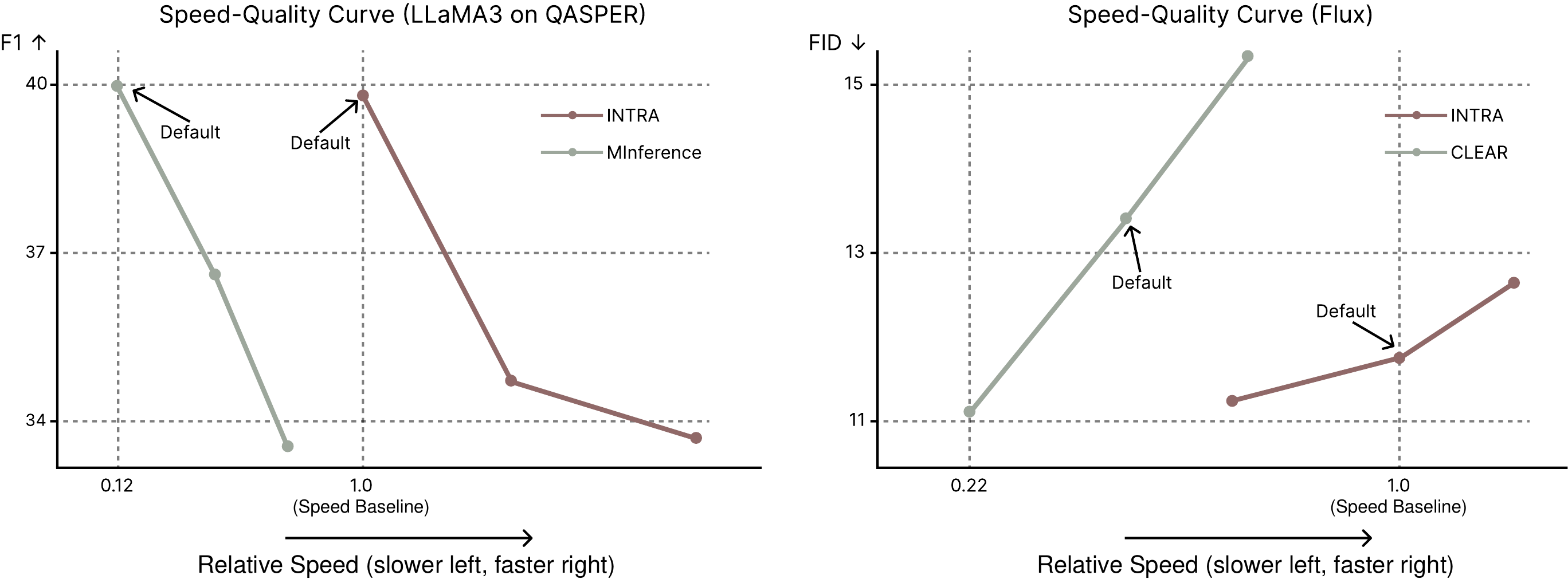}
  \caption{Speed--quality trade-offs for LLaMA 3 on Qasper (left) and \texttt{FLUX.1-dev} (right). Horizontal axes show relative speed with the INTRA default at 1: points nearer the origin are slower, and points farther right are faster.}
  \label{fig:full_speed_quality}
\end{figure}

Figure~\ref{fig:full_speed_quality} varies the group count as INTRA's sparsity control knob: more groups increase sparsity and reduce computation. For LLaMA 3, the group counts are 8, 12, and 16; for \texttt{FLUX.1-dev}, they are 4, 8, and 16. These are numbers of groups, not tokens per group. The curves show the trade-off between relative speed and Qasper F1 or image FID as sparsity changes.

\paragraph{Main-table summary.}
Separately, we average the six task scores reported in Table~\ref{tab:latency_scroll_benchmark}, $\bar{s}_m=\frac{1}{6}\sum_{j=1}^{6}s_{m,j}$. This is an unweighted mean of task-specific metrics in score points. The means are 36.88 for XAttention, 36.44 for MInference, and 36.01 for INTRA. Differences are computed before rounding: INTRA's mean-score gaps are 0.87 and 0.42 points, respectively.

The attention operation summary uses the geometric mean of latency ratios over $\mathcal{L}=\{8\mathrm{K},16\mathrm{K},32\mathrm{K}\}$: $g_B=\bigl(\prod_{L\in\mathcal{L}}t_B(L)/t_{\mathrm{INTRA}}(L)\bigr)^{1/3}$. This gives $2.23\times$ against XAttention ($S=16$) and $14.94\times$ against MInference. These summaries use Table~\ref{tab:latency_scroll_benchmark} directly and are separate from the group-count sweeps in Figure~\ref{fig:full_speed_quality}.

\section{Additional Experiments and Ablations}
\label{appendix:additional_results}
\label{appendix:language_experiments}
This appendix collects image-generation ablations, language adaptation and retrieval studies, implementation costs, and video-generation results.

\subsection{Image Generation Ablations}
\label{appendix:image_ablations}
\label{appendix:inter_freq}
Table~\ref{tab:smaller_dense_model} compares INTRA with Stable Diffusion 1.5 under the same FID protocol. INTRA obtains FID 11.75, compared with 30.02 for the smaller dense diffusion model.

We also ablate how often INTRA switches between Scatter and Gather. The number of layers in Table~\ref{tab:inter_frequency} indicates how many consecutive layers share the same sparse pattern before switching. Interleaving every 2 layers performs best across metrics.

\begin{table}[htbp]
\centering
\caption{Additional image-generation ablations. Left: comparison with a smaller dense diffusion model. Right: ablation on interleaving frequency.}
\vspace{0.5em}
\small
\makebox[\textwidth]{%
  \begin{subtable}[t]{0.34\textwidth}
  \centering
  \caption{Smaller dense model}
  \vspace{1.15em}
  \begin{tabular}{lc}
  \toprule
  \textbf{Method} & \textbf{FID ($\downarrow$)} \\
  \midrule
  SD 1.5 & 30.02 \\
  INTRA & \textbf{11.75} \\
  \bottomrule
  \end{tabular}
  \label{tab:smaller_dense_model}
  \end{subtable}%
  \hfill
  \begin{subtable}[t]{0.58\textwidth}
  \centering
  \caption{Interleaving frequency}
  \begin{tabular}{lccc}
  \toprule
  \textbf{Frequency} & \textbf{FID ($\downarrow$)} & \textbf{LPIPS ($\downarrow$)} & \textbf{CLIPI ($\uparrow$)} \\
  \midrule
  1 layer  & 12.6 & 51.7 & 0.895 \\
  2 layers & \textbf{11.7} & \textbf{49.4} & \textbf{0.904} \\
  3 layers & 13.1 & 51.0 & 0.892 \\
  4 layers & 12.8 & 53.2 & 0.890 \\
  \bottomrule
  \end{tabular}
  \label{tab:inter_frequency}
  \end{subtable}%
}
\end{table}

\subsection{Language Adaptation Protocols}
\label{appendix:llm_finetune}
\label{appendix:retrieval_adaptation}
\paragraph{SCROLLS fine-tuning.}
Table~\ref{tab:llm_no_finetune} compares the original LLaMA 3.1 8B model without fine-tuning, the original model with fine-tuning, and INTRA with fine-tuning. The Qasper result uses LoRA rank 32 and closes most of the gap to the fine-tuned dense model. The earlier language fine-tuning setup was reported as 320 GPU hours. This cost and the rank-32 Qasper setting are separate from the matched retrieval adaptation protocol below.

\begin{table}[h]
\centering
\caption{LLaMA 3.1 8B performance on SCROLLS with and without fine-tuning.}
\vspace{0.5em}
\small
\resizebox{0.85\textwidth}{!}{
\begin{tabular}{lccccc}
\toprule
 & \textbf{GovReport} & \textbf{Qasper} & \textbf{QMSum} & \textbf{QuALITY} & \textbf{SummScreen-FD} \\
\midrule
Original model (no fine-tune) & 24.13 & 34.91 & 21.17 & 0.0 & 16.77 \\
Original model & \textbf{31.05} & \textbf{40.09} & \textbf{25.37} & 26.0 & \textbf{20.51} \\
INTRA & 27.77 & 39.64 & 24.45 & \textbf{29.0} & 19.71 \\
\bottomrule
\end{tabular}}
\label{tab:llm_no_finetune}
\end{table}

\paragraph{Matched retrieval adaptation.}
For the NIAH comparisons and language ablations, all methods, including the dense control, use the same data and adaptation budget. The training mixture contains 60\% public train-split conversations in the evaluation format, 25\% synthetic retrieval examples, and 15\% plain continuation. The data are packed into 400 blocks of 16,384 tokens and are disjoint from all evaluation documents. We train for 1,200 steps using rank-16 LoRA on the attention $q/k/v/o$ and MLP gate/up/down projections. Supervision comes from each model's own dense outputs, following knowledge distillation~\citep{distill}, without ground-truth task labels. One adapter is trained per model and reused across its evaluation tasks. For LLaMA-3.1-8B, the adapters contain 41.9M trainable parameters (approximately 0.5\% of the model), and each run takes about one hour on eight A100 GPUs, or approximately 8 GPU hours. Qwen3-8B uses the same adaptation recipe and budget, with the number and retained density of its global sparse layers configured as described in the main text.

\subsection{Window-Width Ablation}
\label{appendix:window_width}
We further evaluate LLaMA-3.1-8B models using sliding-window attention in all 32 layers under the same adaptation budget. Increasing the window width from 575 to 4096 tokens raises the NIAH score at 4K from 0.111 to 1.000, whereas the score at 32K increases only from 0.000 to 0.111. These results show that wider windows alone remain insufficient for reliable retrieval at 32K under the evaluated protocol. Table~\ref{tab:window_width_ablation} reports the measured scores.

\begin{table}[htbp]
\centering
\caption{Window-width ablation on LLaMA-3.1-8B with sliding-window attention in all 32 layers. Scores are measured NIAH results at 4K and 32K context lengths.}
\label{tab:window_width_ablation}
\small
\begin{tabular}{lcc}
\toprule
\textbf{Window width} & \textbf{NIAH at 4K $\uparrow$} & \textbf{NIAH at 32K $\uparrow$} \\
\midrule
575 & 0.111 & 0.000 \\
2048 & 0.556 & 0.000 \\
4096 & 1.000 & 0.111 \\
\bottomrule
\end{tabular}
\end{table}

\subsection{Attention-Configuration Comparison}
\label{appendix:language_configuration}
Table~\ref{tab:language_configuration} compares alternative attention configurations of LLaMA-3.1-8B under a fixed adaptation budget. The configuration with sparse attention in half the layers retains a density of 0.125 within each selected layer, while the four-layer configuration retains 0.5 within each selected layer. The four-layer configuration achieves the highest NIAH score among the sparse configurations shown: 0.923 at overall density 0.123, compared with 0.222 at density 0.234 for wider windows. Across these density and layer-allocation choices, concentrating global attention in four layers provides the strongest retrieval result among the tested sparse configurations.

\begin{table}[htbp]
\centering
\caption{Language attention configurations under the same adaptation budget. Retained density refers to the overall attention configuration, normalized to dense attention at 1.000.}
\label{tab:language_configuration}
\small
\begin{tabular}{p{0.50\textwidth}cc}
\toprule
\textbf{Configuration} & \textbf{Retained density} & \textbf{Mean NIAH $\uparrow$} \\
\midrule
Sparse attention in 16 layers (per-layer density 0.125) & 0.098 & 0.130 \\
Wider windows & 0.234 & 0.222 \\
Four global sparse layers (per-layer density 0.5) & 0.123 & 0.923 \\
Dense + LoRA & 1.000 & 1.000 \\
\bottomrule
\end{tabular}
\end{table}

\subsection{LoRA Adaptation Ablation}
\label{appendix:no_lora}
Without LoRA adaptation, NIAH scores range from 0.02 to 0.09 across the sparse methods and configurations tested in this study. These unadapted runs provide the reference for measuring the effect of LoRA adaptation on retrieval.

\subsection{Mask Construction and Caching}
\label{appendix:mask_construction}
For the block-level language configuration, the block mask is constructed once per configuration and cached for reuse. Mask construction takes 3.4\,ms at a context length of 16K. This is a one-time mask-construction cost; repeated calls with the same cached configuration reuse the mask.

\subsection{Attention Timing Comparison with MoBA and NSA}
\label{appendix:moba_speed}
Table~\ref{tab:moba_relative_speed} compares INTRA, MoBA~\citep{moba}, and NSA~\citep{nsa} on one A100-SXM4-80GB, measured sequentially with LLaMA-3.1-8B attention shapes: 32 query heads, eight key/value heads (GQA-4), head dimension 128, bfloat16, batch size one, and causal prefill. We time attention only, taking the median of 50 repetitions for each layer. Panel (a) sums these medians across 32 layers; panel (b) reports single-layer latency. For method $A$ and reference $B$, speedup is $t_B/t_A$.

\paragraph{Configurations and density.}
The fixed \texttt{g2s4} schedule uses four Scatter layers with $G=2$ and 28 sliding-window layers with width 576. Its window contribution to overall retained density varies with context length. The separate single-layer comparison uses Scatter with $G=8$ and a target retained density of 0.125. MoBA uses the official implementation, selecting two of 16 blocks (density 0.125), with the fastest tested chunk setting at each length. NSA's paper setting is $(l,d,l',n,w)=(32,16,64,16,512)$; its measured densities at 4K, 16K, and 64K are 0.578, 0.220, and 0.104, respectively. For density-constrained NSA, we sweep 18 combinations of $(l,d)$, $l'$, and $w$, binary-search the largest $n$ satisfying density $\leq 0.125$, and select the fastest configuration. The selected $(l,d,l',n,w)$ tuples are $(64,32,64,3,128)$ at 4K, $(64,32,64,6,512)$ at 16K, and $(128,64,128,27,512)$ at 64K.

\paragraph{Reading the comparison.}
We label density-constrained NSA as \emph{NSA (tuned, $\rho\leq0.125$)}, where $\rho$ denotes retained attention density. The 0.125 cap constrains connectivity; the speedup columns report $t_{\mathrm{baseline}}/t_{\mathrm{INTRA}}$. For example, at 16K, $569.5/59.7\approx9.5\times$ corresponds to an 89.5\% reduction in the measured attention time. Table~\ref{tab:niah_attention_timing} brings the 32-layer comparison into the main text; Table~\ref{tab:moba_relative_speed} additionally reports NSA's paper configuration and the separate single-layer experiment.

\begin{table}[htbp]
\centering
\caption{Causal prefill attention latency (ms) on A100-SXM4-80GB. (a) Sum of 32 per-layer medians for the fixed INTRA \texttt{g2s4} schedule and baselines. (b) Single-layer comparison with target density 0.125. NSA (tuned) is the fastest tested setting satisfying $\rho\leq0.125$; NSA (paper) uses its paper hyperparameters. INTRA speedup over a baseline is $t_{\mathrm{baseline}}/t_{\mathrm{INTRA}}$.}
\label{tab:moba_relative_speed}
\small
\setlength{\tabcolsep}{3pt}
\begin{subtable}{\textwidth}
\centering
\caption{32-layer attention totals; lower latency is better.}
\begin{tabular}{lrrrrrrr}
\toprule
\multicolumn{6}{c}{\textbf{Attention latency (ms $\downarrow$)}} & \multicolumn{2}{c}{\textbf{INTRA speedup over}} \\
\cmidrule(lr){1-6}\cmidrule(lr){7-8}
\textbf{Context} & \shortstack{\textbf{Dense}\\(SDPA)} & \shortstack{\textbf{INTRA}\\(\texttt{g2s4})} & \textbf{MoBA} & \shortstack{\textbf{NSA}\\(paper)} & \shortstack{\textbf{NSA (tuned)}\\$\rho\leq0.125$} & \textbf{MoBA} & \shortstack{\textbf{NSA}\\(tuned)} \\
\midrule
4K & 27.4 & \textbf{11.3} & 95.2 & 238.8 & 117.7 & $8.4\times$ & $10.4\times$ \\
16K & 381.0 & \textbf{59.7} & 349.7 & 936.3 & 569.5 & $5.9\times$ & $9.5\times$ \\
64K & 6043.3 & \textbf{577.2} & 2073.0 & 4390.6 & 4235.6 & $3.6\times$ & $7.3\times$ \\
\bottomrule
\end{tabular}
\end{subtable}
\par\medskip
\begin{subtable}{\textwidth}
\centering
\caption{Single-layer latency with Scatter $G=8$.}
\begin{tabular}{lrrrrr}
\toprule
\textbf{Context} & \textbf{Dense} & \textbf{INTRA Scatter} & \textbf{MoBA} & \shortstack{\textbf{NSA (tuned)}\\$\rho\leq0.125$} & \shortstack{\textbf{INTRA speedup}\\\textbf{over NSA (tuned)}} \\
\midrule
4K & 0.856 & \textbf{0.223} & 2.976 & 3.677 & $16.5\times$ \\
16K & 11.906 & \textbf{1.955} & 10.927 & 17.797 & $9.1\times$ \\
64K & 188.852 & \textbf{28.541} & 64.782 & 132.363 & $4.6\times$ \\
\bottomrule
\end{tabular}
\end{subtable}
\end{table}

\paragraph{Comparison.}
INTRA has the lowest attention latency in every reported setting. Across 4K--64K, the fixed \texttt{g2s4} schedule is $3.6$--$8.4\times$ faster than MoBA and $7.3$--$10.4\times$ faster than density-constrained NSA, reaching $10.47\times$ the dense attention speed at 64K. The single-layer comparison further isolates the Scatter kernel: at target density 0.125, it is $3.84\times$, $6.09\times$, and $6.62\times$ faster than dense attention at 4K, 16K, and 64K, and $4.6$--$16.5\times$ faster than density-constrained NSA. The advantage thus persists both for a sparse-layer/window schedule and for a single global sparse layer. INTRA's static indexing avoids input-dependent selection and routing, providing an efficient execution path for the evaluated workloads. These timing measurements complement the separate matched-adaptation retrieval comparison in Table~\ref{tab:niah_retrieval}.

\subsection{Video Generation Results}
\label{appendix:wan_vbench}

Table~\ref{tab:wan_vbench_selected} reports all 13 evaluated VBench dimensions at $480\times832$ with 49 frames, covering temporal consistency, perceptual quality, style, and object and spatial criteria. The summary in the main text is the unweighted arithmetic mean of these 13 per-dimension scores. We report both dense and INTRA scores together with their absolute differences.

\begin{table}[!htb]
\centering
\caption{All 13 evaluated VBench~\citep{vbench} dimensions on Wan2.2-TI2V-5B~\citep{wan,wan22}, using 944 prompts and five samples per prompt. Higher scores are better; the final column reports absolute dense--INTRA differences.}
\label{tab:wan_vbench_selected}
\small
\begin{tabular}{lccc}
\toprule
\textbf{VBench dimension} & \textbf{Dense} & \textbf{INTRA} & \textbf{Absolute difference} \\
\midrule
Subject consistency & 0.9427 & 0.9405 & 0.0022 \\
Background consistency & 0.9587 & 0.9551 & 0.0036 \\
Aesthetic quality & 0.5362 & 0.5168 & 0.0194 \\
Imaging quality & 0.6036 & 0.5472 & 0.0564 \\
Temporal flickering & 0.9886 & 0.9874 & 0.0012 \\
Motion smoothness & 0.9664 & 0.9689 & 0.0025 \\
Dynamic degree & 0.4583 & 0.4667 & 0.0084 \\
Overall consistency & 0.2435 & 0.2400 & 0.0035 \\
Appearance style & 0.2277 & 0.2222 & 0.0055 \\
Temporal style & 0.2427 & 0.2356 & 0.0071 \\
Object class & 0.0000 & 0.0000 & 0.0000 \\
Multiple objects & 0.0000 & 0.0000 & 0.0000 \\
Spatial relationship & 0.0000 & 0.0000 & 0.0000 \\
\bottomrule
\end{tabular}
\end{table}

\paragraph{Video latency and attention scaling.}
Table~\ref{tab:wan_timing} measures Wan2.2-TI2V-5B on A100-SXM4-80GB at three sequence lengths. End-to-end timing includes text encoding, 50 denoising steps with two classifier-free-guidance evaluations per step (100 DiT forwards), and VAE decoding. Standalone forward timing uses batch size one and all 30 DiT blocks; self-attention timing sums the 30 self-attention layers. Pipeline DiT totals and standalone forward times come from separate measurements.

\begin{table}[!htb]
\centering
\caption{Wan2.2-TI2V-5B video timing on A100-SXM4-80GB. Arrows show dense $\rightarrow$ INTRA; speedups are dense time divided by INTRA time. End-to-end and pipeline components are in seconds, while standalone forward and self-attention times are in milliseconds. The 49-frame setting matches VBench; 121 frames corresponds to the native five-second setting at 24 fps. The 241-frame setting measures standalone speed only; dashes denote unmeasured pipeline timings.}
\label{tab:wan_timing}
\small
\begin{tabular}{lccc}
\toprule
\textbf{Resolution} & $480\times832$ & $704\times1280$ & $704\times1280$ \\
\textbf{Frames / video tokens} & 49 / 5,070 & 121 / 27,280 & 241 / 53,680 \\
\midrule
\multicolumn{4}{l}{\textit{End-to-end generation and pipeline components (s)}} \\
Total generation & $44.1\rightarrow41.2$ & $340.9\rightarrow238.4$ & --- \\
End-to-end speedup & $1.07\times$ & $\mathbf{1.43\times}$ & --- \\
DiT total (100 forwards) & $40.9\rightarrow38.0$ & $326.9\rightarrow224.2$ & --- \\
VAE decoding & 2.6 & 13.4 & --- \\
\midrule
\multicolumn{4}{l}{\textit{Standalone DiT forward: batch size one, all 30 blocks}} \\
DiT forward (ms) & $399\rightarrow371$ & $3257\rightarrow2198$ & $9143\rightarrow4855$ \\
Forward speedup & $1.07\times$ & $1.48\times$ & $\mathbf{1.88\times}$ \\
Self-attention (ms) & $49.3\rightarrow23.2$ & $1423\rightarrow342$ & $5528\rightarrow1129$ \\
Self-attention speedup & $2.12\times$ & $4.16\times$ & $\mathbf{4.90\times}$ \\
Attention share of dense forward & $12\%$ & $44\%$ & $60\%$ \\
\bottomrule
\end{tabular}
\end{table}

At 5,070 tokens, self-attention accounts for 12\% of dense forward time; accelerating it by $2.12\times$ yields a $1.07\times$ end-to-end speedup. At 27,280 tokens, attention's share rises to 44\%, and its $4.16\times$ speedup translates into $1.48\times$ faster standalone DiT execution and $1.43\times$ faster generation, saving 102.5 seconds per video. At 53,680 tokens, attention occupies 60\% of dense forward time: INTRA reduces self-attention from 5528 to 1129\,ms and the complete forward from 9143 to 4855\,ms. The increasing attention share explains why structured sparsity gives larger overall gains as video sequence length grows.

\clearpage
\section{Qualitative Results}
\label{appendix:visual_compare}
\begin{figure}[htbp]
    \centering
    \includegraphics[width=1\textwidth, trim=0 113 0 120, clip]{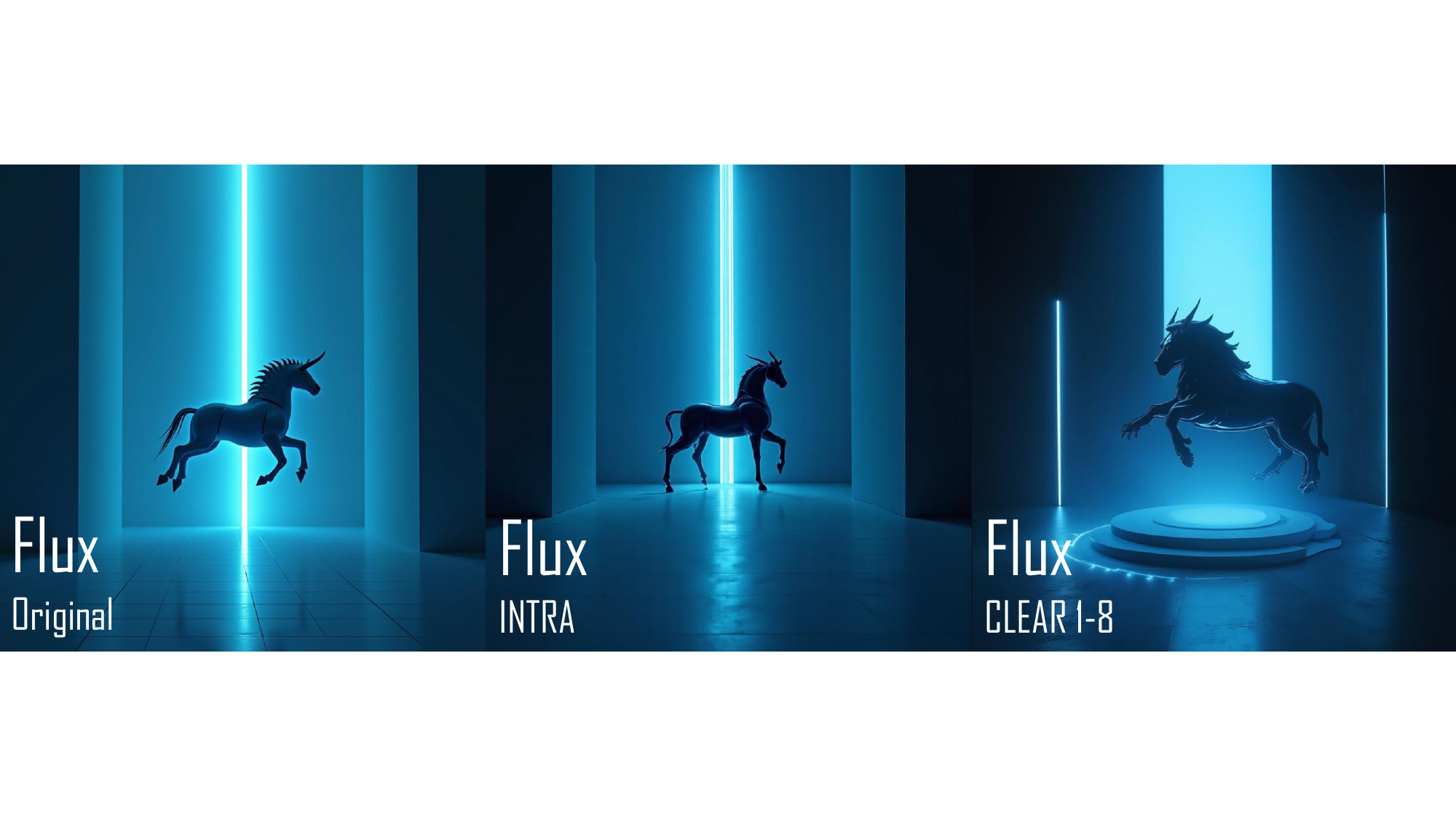}
    \caption{\texttt{FLUX.1-dev} outputs: original FlashAttention 2, INTRA, and CLEAR (left to right).}
    \label{fig:enter-label}
\end{figure}

\begin{figure}[htbp]
  \centering
  \includegraphics[width=\textwidth,height=0.78\textheight,keepaspectratio]{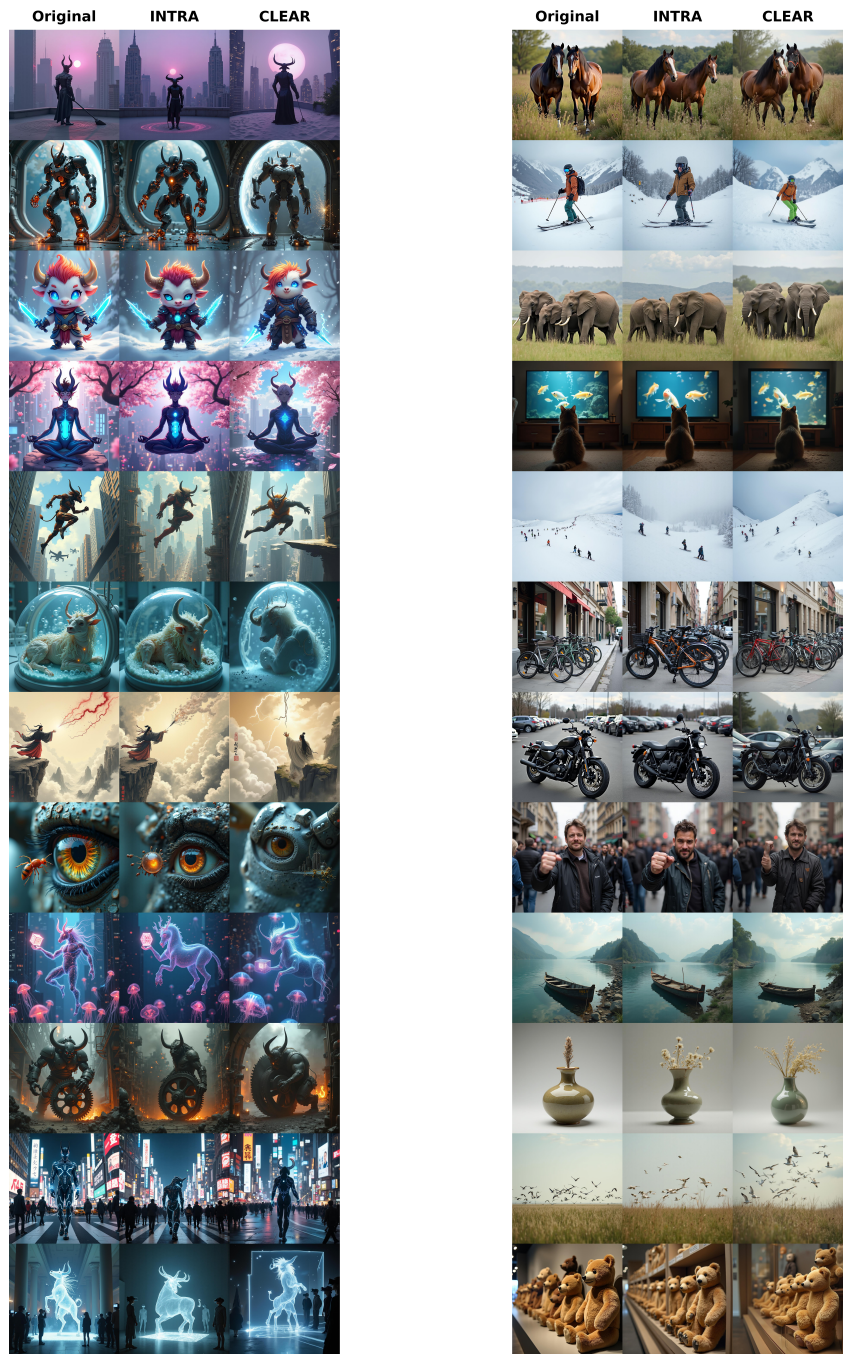}
  \caption{Visual examples comparing original \texttt{FLUX.1-dev}, INTRA, and CLEAR.}
  \label{fig:qualitative1}
\end{figure}

\begin{figure}[htbp]
  \centering
  \includegraphics[width=\textwidth,height=0.82\textheight,keepaspectratio]{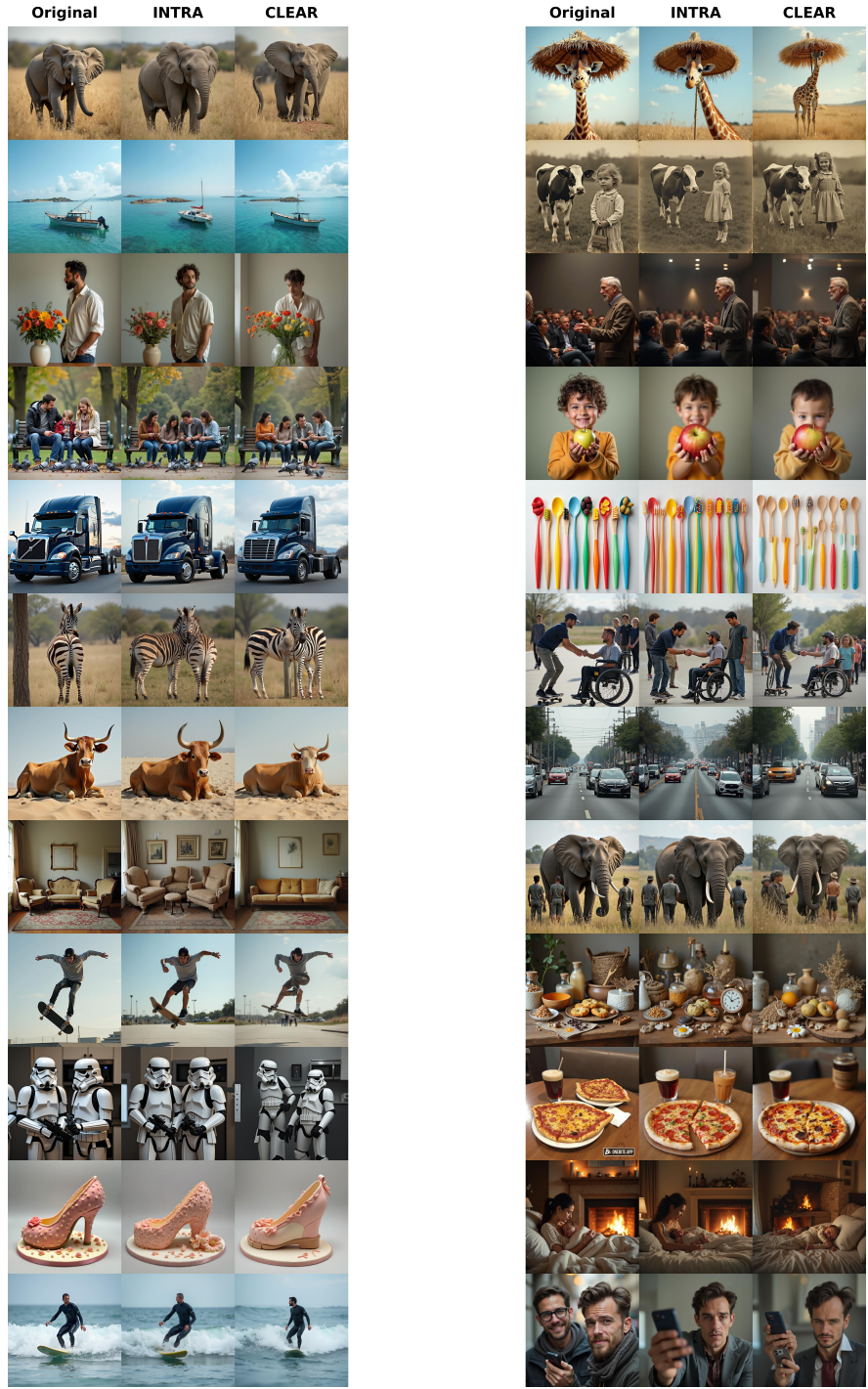}
  \caption{Additional visual examples with the same column order as Figure~\ref{fig:qualitative1}: original \texttt{FLUX.1-dev}, INTRA, and CLEAR within each triplet.}
  \label{fig:qualitative2}
\end{figure}

Figure~\ref{fig:enter-label} provides an overview; Figures~\ref{fig:qualitative1} and~\ref{fig:qualitative2} provide additional visual comparisons between the original \texttt{FLUX.1-dev} model, INTRA, and CLEAR.

\clearpage

\end{document}